\documentclass[runningheads]{llncs}

\usepackage{eccv}

\usepackage{eccvabbrv}

\usepackage{graphicx}
\usepackage{booktabs}
\usepackage{color}

\definecolor{turquoise}{cmyk}{0.65,0,0.1,0.3}
\definecolor{purple}{rgb}{0.65,0,0.65}
\definecolor{dark_green}{rgb}{0, 0.5, 0}
\definecolor{orange}{rgb}{0.8, 0.6, 0.2}
\definecolor{red}{rgb}{0.8, 0.2, 0.2}
\definecolor{darkred}{rgb}{0.6, 0.1, 0.05}
\definecolor{blueish}{rgb}{0.0, 0.3, .6}
\definecolor{light_gray}{rgb}{0.7, 0.7, .7}
\definecolor{pink}{rgb}{1, 0, 1}
\definecolor{greyblue}{rgb}{0.25, 0.25, 1}
\definecolor{tab_blue}{HTML}{1f77b4}
\definecolor{tab_orange}{HTML}{ff7f0e}
\definecolor{LightRed}{rgb}{0.99,0.89,0.89}
\definecolor{mesh_misty_rose}{HTML}{e6aaa3}
\definecolor{mesh_yellow}{HTML}{ffba00}
\definecolor{MyDarkBlue}{rgb}{0.02,0.02,0.6}

\usepackage[accsupp]{axessibility}  

\newcommand{\pipelinename}{\textit{ZeST}\xspace}

\usepackage{hyperref}

\usepackage{orcidlink}

\begin{document}


\title{\textit{ZeST}: Zero-Shot Material Transfer \\from a Single Image}

\titlerunning{ZeST}

\author{Ta-Ying Cheng\inst{1}$^,$\inst{2}\and
Prafull Sharma\inst{3} \and
Andrew Markham\inst{1} \and \\Niki Trigoni\inst{1} \and Varun Jampani\inst{2}}

\authorrunning{Cheng et al.}

\institute{University of Oxford \and
Stability AI \and
MIT CSAIL
}

\maketitle
\vspace{-1mm}
\begin{figure}[h]
    \vspace{-2em}
    \centering
    \includegraphics[width=\columnwidth]{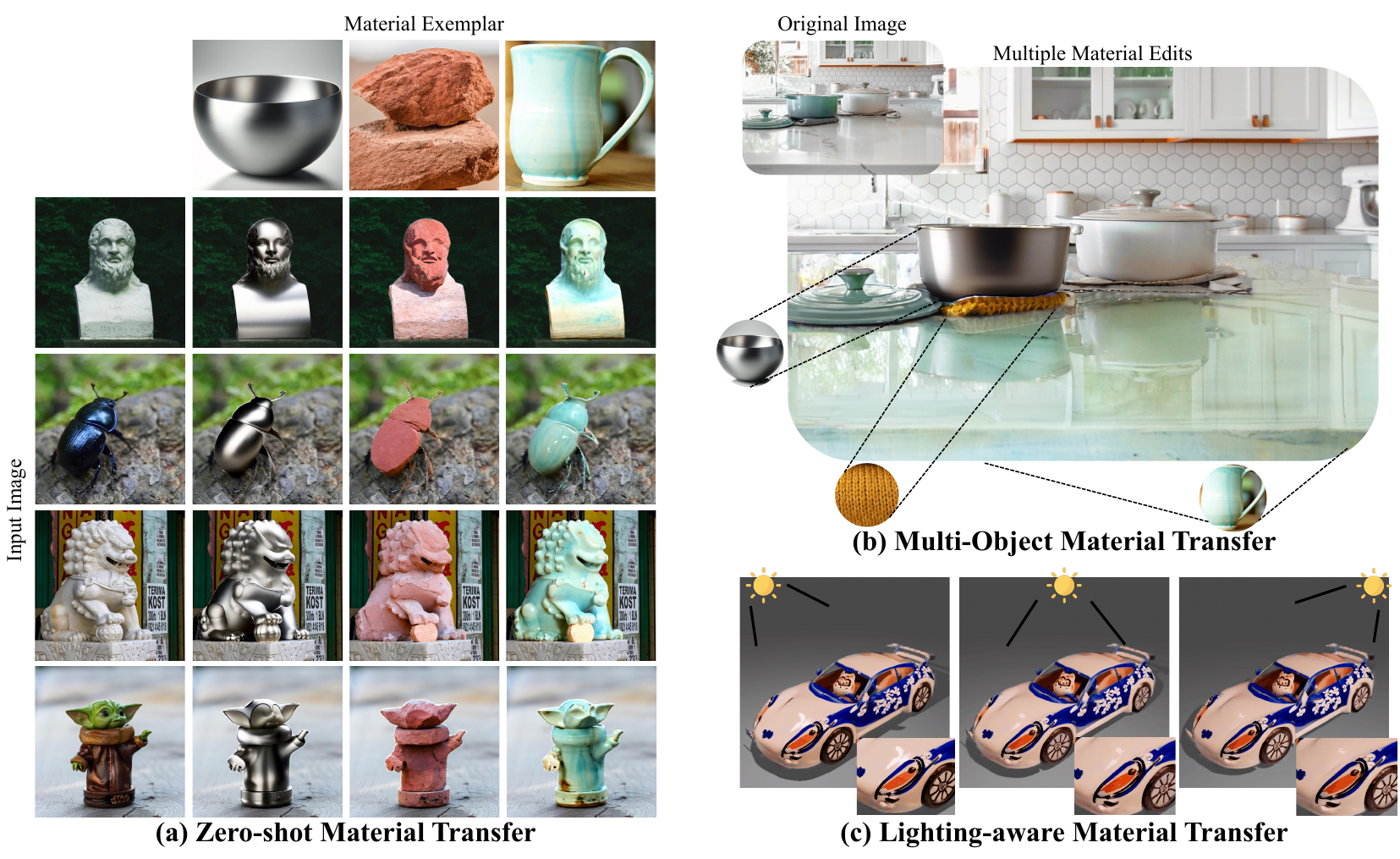}
    \caption{\textbf{Overview.} We present \pipelinename, a zero-shot single-image approach to \textbf{(a)} transfer material from an examplar image to an object in the input image. \textbf{(b)} \pipelinename can easily be extended to perform multiple material edits in an single image, and \textbf{(c)} perform implicit lighting-aware edits on rendering of a textured mesh.}
    \vspace{-3em}
    \label{fig:teaser}
\end{figure}
\begin{abstract}
We propose \pipelinename, a method for zero-shot material transfer to an object in the input image given a material exemplar image.
\pipelinename leverages existing diffusion adapters to extract implicit material representation from the exemplar image. This representation is used to transfer the material using pre-trained inpainting diffusion model on the object in the input image using depth estimates as geometry cue and grayscale object shading as illumination cues. The method works on real images without any training resulting a zero-shot approach.
Both qualitative and quantitative results on real and synthetic datasets demonstrate that \pipelinename
outputs photorealistic images with transferred materials.  
We also show the application of \pipelinename to perform multiple edits and robust material assignment under different illuminations. 

Project Page: \href{https://ttchengab.github.io/zest/}{https://ttchengab.github.io/zest}

\end{abstract}

\section{Introduction}\label{sec:intro}

Editing object materials in images (\textit{e.g.,} changing a marble statue into a steel statue) is useful for several graphics and design applications such as game design, e-commerce, etc. It is a highly challenging and time-consuming task even for expert artists and graphic designers -- typically requires explicit 3D geometry and illumination estimation followed by careful tuning of the target material properties (\textit{e.g.,} metallic, roughness, transparency). Previous works try to alleviate the tedious material specification by synthesizing textures given input text prompts \cite{yeh2024texturedreamer,richardson2023texture}. However, they are focused on texturing 3D meshes, which overlooks some of the unique challenges for material editing in 2D images, such as illumination. Another work~\cite{sharma2023alchemist} proposes fine-grained material editing on images, but it cannot directly transfer materials from a given exemplar.

In this work, we aim to make 2D-to-2D material editing practical by eliminating the need for any 3D objects as well as explicit specification of material properties. Given a single image of an object and another material exemplar image, our goal is to transfer the material appearance from the exemplar to the target object directly in 2D. See Fig.~\ref{fig:teaser} for some sample input and material exemplar images. We do not assume any access to the ground-truth 3D shapes, illumination, or even the material properties, making this problem setting practical and widely applicable for material editing.

This setup is particularly challenging from two perspectives. First, an explicit approach to material transfer requires an understanding of many object-level properties in both the exemplar and the input image, such as geometry and illumination. Subsequently, we have to disentangle the material information from these properties and apply it to the new image; the entire process has several unsolved components. Second, there currently exists no real-world datasets for supervising this task. Collecting high-quality datasets presenting the same object with multiple materials and exemplars may be quite tedious.

One of the main contributions of this work in alleviating these challenges is a zero-shot approach that can implicitly transfer arbitrary material appearances from a given 2D exemplar image onto a target 2D object image, without explicitly estimating any 3D or material properties from either image. We call our approach `\pipelinename', as it does not require multiple exemplars or any training like previous works, making it easy to generalize to any images in the wild.

With \pipelinename, we propose a carefully designed pipeline that repurposes several recent advances in 2D image generation and editing for our problem setting. At a high level, we adapt the geometry-guided generation (\textit{e.g.,} ControlNet~\cite{zhang2023adding}) and also exemplar-guided generation (\textit{e.g.,} IP-Adapter~\cite{ye2023ip}) to implicitly isolate and transfer material appearance from a source exemplar to the target image while applying a foreground decolored image and inpainting for illumination cues. Our key contribution is presenting a simple pipeline with careful design choices that can be used to tackle a highly challenging problem of 2D-to-2D material transfer.

Since this is a new problem setting, we created both synthetic and real-world evaluation datasets with material exemplars and object images.
Extensive qualitative and quantitative evaluations demonstrate that \pipelinename excels in photo-realism and material accuracy in the output images when compared against various baselines while being completely training-free. See Fig.~\ref{fig:teaser}(a) for sample results of~\pipelinename.
With our pipeline, artists can grab pre-designed materials as material exemplars and directly transfer them to real-world images. By using different object masks, we can also use \pipelinename to cast different materials to multiple objects present in a single image (Fig.~\ref{fig:teaser} (b)). In addition, with slight alteration of the inputs,~\pipelinename can perform light-aware material transfer by changing the reflections while keeping textural patterns consistent (Fig.~\ref{fig:teaser} (c)); this method can have potential application when used in conjunction with 3D texture generation methods~\cite{chen2023text2tex}. 

In summary,~\pipelinename has several favorable properties for material editing:
\vspace{-3mm}
\begin{itemize}
    \item[$\circ$] \textbf{Zero-shot, training free, single-image material transfer.} By leveraging 2D generative priors,~\pipelinename works in a zero-shot manner without needing dataset finetuning. Unlike some contemporary works~\cite{yeh2024texturedreamer} that implicitly capture material properties using several material images,~\pipelinename only needs a single material exemplar image to transfer the material in pixel space.
    \item[$\circ$] \textbf{No explicit 3D, illumination or materials.} With 2D depth and segmentation estimation (which are readily available these days) and implicit material transfer, we eliminate the need for explicit specification of 3D meshes, illumination or material properties (say, in terms of BRDF).
    \item[$\circ$] \textbf{Several downstream applications.} Given the simplistic and practical nature of our approach,~\pipelinename can be used for several downstream graphics applications such as applying pre-designed materials to real-world images, editing multiple object materials in a single image, and perform lighting-aware material transfer given untextured mesh renderings.
\end{itemize}

\section{Related Work}

\textbf{Diffusion Models.} Denoising Diffusion Probabilistic models have emerged as the state-of-the-art for class-conditional and text-prompt conditioned image generation~\cite{dhariwal2021diffusion,ho2020denoising,ho2022cascaded,ho2022classifier,song2019generative,karras2022elucidating,kang2023scaling}. These models generate photorealistic images with exemplary geometry, materials, illumination, and scene composition. The models have been extended to be conditioned on input images for computational photography tasks such as super-resolution, style transfer, and inpainting. 

Further work demonstrate controllable generation conditioned on text-based instructions~\cite{hertz2022prompt,voynov2023p+,ge2023expressive,cao2023masactrl}, semantic segmentation~\cite{bar2023multidiffusion}, bounding box~\cite{li2023gligen,chen2024training,yang2023reco,wang2024instancediffusion}, depth~\cite{zhao2024uni,bhat2023loosecontrol}, sketch~\cite{zhang2023adding,mou2023t2i}, and image prompt~\cite{ye2023ip}. Prompt-to-prompt and Prompt+ edit the input image by performing inversion followed by the introduction of new terms and reweighting the effect of terms in the input prompt~\cite{hertz2022prompt,voynov2023p+}. InstructPix2Pix performs edits an input image conditioned on an instruction~\cite{brooks2023instructpix2pix}. Ge et al. proposed rich text based image editing allowing for style assignment and specific description to specific terms in the prompt~\cite{ge2023expressive}. While these methods edit the image semantically and high-level descriptions, assigning specific materials using text-based approach is challenging since text acts as a limiting modality for describing textures. 

A collection of reference images can be used to learn concepts which can be further included in text prompts to generate images with the learned concepts~\cite{ruiz2022dreambooth,chen2024subject,kumari2023multi}. Spatial modalities such as depth and sketches have been used for controlling the generated images~\cite{zhang2023adding,mou2023t2i,ye2023ip}. Pre-trained text-to-image models can be leveraged for 3D-aware image editing using language and depth cues~\cite{cheng2024learning,michel2024object,pandey2023diffusion}. The use of ControlNet has been extended by Bhat et al. to use depth for controlling the scene composition while maintaining other scene attributes~\cite{bhat2023loosecontrol}. Object orientation, illumination, and other object attributes can be controlled in a continuous manner using ControlNet and learned continuous tokens embedding the 3D properties~\cite{cheng2024learning}.

\noindent\textbf{Material acquisition and editing.} Material acquisition and editing is an active field of research taking into account illumination and object geometry. 
Previous work has demonstrated material acquisition under known illumination conditions and camera~\cite{aittala2013practical,aittala2015two,deschaintre2019flexible}. Such acquisition in the wild requires localizing objects with similar materials, which has been facilitated by supervised material segmentation and leveraging pre-trained vision representation backbones~\cite{bell2015material,liang2022multimodal,upchurch2022dense,sharma2023materialistic}.
Khan et al. introduced in-image material editing using estimates of depth~\cite{khan2006image}. Recent works have employed generative adversarial networks~\cite{goodfellow2020generative} for perceptual material editing~\cite{subias2023wild,delanoy2022generative} and physical shader-based editing using text-to-image models~\cite{sharma2023alchemist}. The use of generative models has been extended to explicitly learning materials \cite{lopes2023material} and texturing 3D meshes~\cite{yeh2024texturedreamer,chen2023text2tex,richardson2023texture,cao2023texfusion}.

In our work, we aim to use pre-trained image generation diffusion models to perform exemplar-based material transfer from a single image. We aim to use ControlNet and IP-adapter to perform material transfer in a zero-shot way without any training.

\section{Method}

In this section, we describe our method \pipelinename that performs exemplar-based material transfer.
Recent methods perform the related problem of texture synthesis on meshes ~\cite{richardson2023texture,yeh2024texturedreamer} by finetuning a diffusion model on 3-5 material exemplar images to capture the texture/material in the latent space.
On the contrary,~\pipelinename only requires a single material exemplar image and a single input image, accomplishing material transfer in a zero-shot, training-free manner.

\begin{figure}[t]
    \centering
    \includegraphics[width=\textwidth]{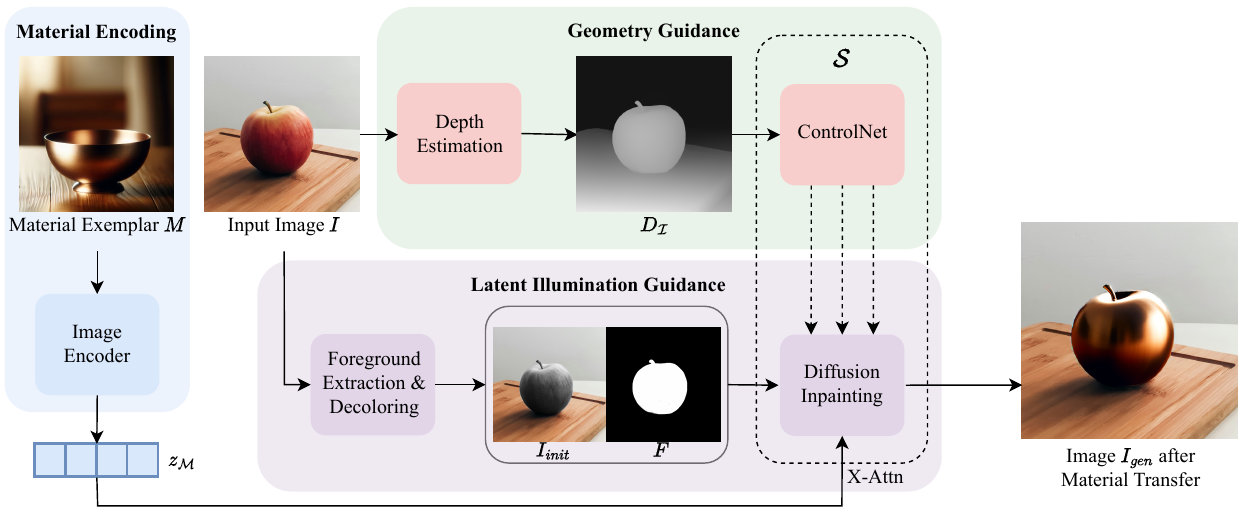}
    \caption{\textbf{\pipelinename Architecture.} Given a material exemplar $M$ and an input image $I$, we first encode material exemplar with an image encoder (\textit{e.g.,} IP-Adaptor). Concurrently, we convert the input image into a depth map $D_I$ and a foreground-grayscaled image $I_{init}$to feed into the geometry and latent illumination guidance branch, respectively. By combining the two sources of guidance with the latent features from the material encoding, \pipelinename can transfer the material properties onto the object in input image while preserving all other attributes.
    } 
    \vspace{-1em}
    \label{fig:method}
\end{figure}
\subsection{Problem Setting}
Given a material exemplar image $M$ and an input image $I$, we aim to output an edited image $I_{gen}$ from $I$ by transferring the material from the material exemplar to the object in the input image while preserving other object and scene properties (e.g. object geometry, background, lighting etc.). 
Performing this task requires understanding the material, geometry, and illumination from both the exemplar and the input image.

In practice, estimating all the aforementioned object-level properties and further isolating material information explicitly from $M$ is challenging since these properties are entangled in the pixel space. Therefore, we propose to tackle this problem in the latent space of diffusion models. Specifically, we aim to extract a latent representation $z_M$ containing the material and texture information that we can then inject into a generative diffusion model $\mathcal{S}$ to generate $I_{gen}$.

\subsection{\pipelinename Overview}
Since there currently exists no synthetic/real image dataset to supervise the learning of a 2D-to-2D material transfer, we perform the material transfer in a zero-shot training-free manner. Therefore, we first break down this complex task into sub-problems of (1) encoding the material exemplar, (2) geometry-guided image editing,
and (3) making the generation process illumination-aware. Given the recent advances in high-fidelity diffusion models and complementary adapters for image generation, we leverage existing pre-trained modules to tackle each of the sub-problems that together compose our pipeline to perform image-prompted material editing.

Figure~\ref{fig:method} presents an overview of our pipeline, which comprises three branches to guide the material, geometry, and lighting information, respectively. The Material Encoding branch takes the material exemplar image $M$ as input, which is processed by the image encoder to obtain a material latent representation $z_M$.

Concurrently, we feed the input image $I$ into Geometry Guidance and Latent Illumination Guidance Branch. The Geometry Guidance branch computes the depth map $D_I$ for the image $I$, which is used as the input to ControlNet. The Latent Illumination Guidance branch computes a foreground mask $F$ using $I$ and creates a foreground-grayscale image $I_{init}$, which we use as input to the Diffusion Inpainting pipeline. We concatenate the embeddings from ControlNet with the inpainting diffusion model at the corresponding and inject the material embedding $z_M$ through the cross-attention. The output of the inpainting diffusion model, $I_{gen}$, with the edited image containing the object in $I$ cast with material from exemplar image $M$.

Our design choices to facilitate computation of material embedding, geometry guidance, and illumination cues are discussed in the following sections.

\subsection{Encoding Material Exemplar}
Given the material exemplar image $M$, this branch encodes the image into a latent representation while preserving its material properties. Previous works~\cite{richardson2023texture,yeh2024texturedreamer} address this by finetuning a text-to-image diffusion model to encode the image into a rare token, implicitly treating the rare token as a latent representation that can be used in conjunction with other texts for image generation. However, this approach of optimizing for the material token requires the time-consuming step for every new material exemplar and usually requires 3-5 images to prevent overfitting.

\begin{figure}[t]
    \centering
    \includegraphics[width=\columnwidth]{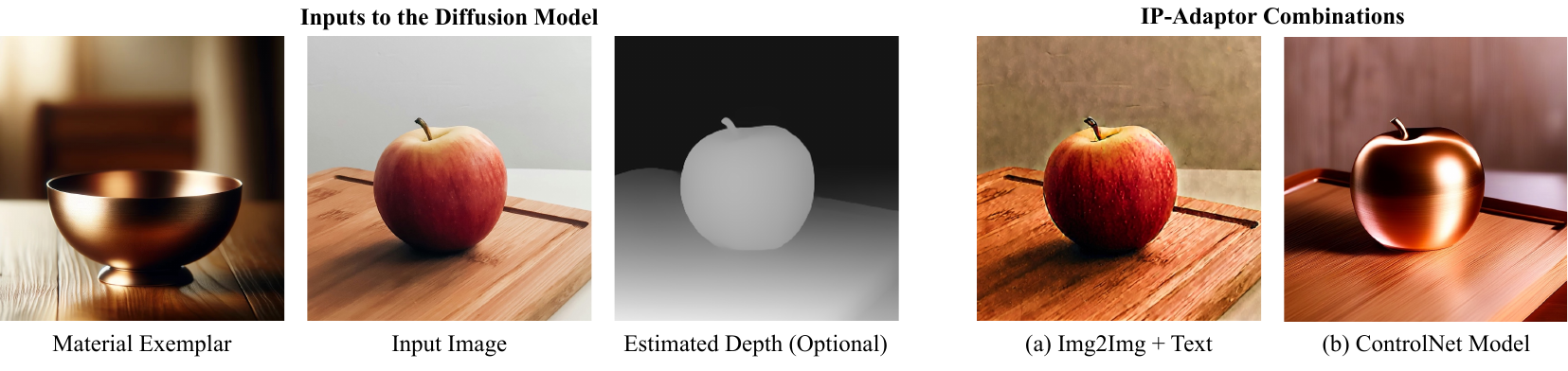}
    \caption{\textbf{The design choice of IP-Adaptor with ControlNet}. Given the material exemplar and the input image, we dive into the different choices of utilizing the IP-Adaptor. In particular we realize that an Img2Img + text module (a) wouldn't properly transfer the materials properly to the main object. On the other hand, ControlNet (b) will preserve the geometry information of the given input. We thus utilize this as the starting point for geometry guidance to further explore the best illumination cues.}
    \vspace{-1em}
    \label{fig:ip_adaptor_ablation}
\end{figure}
We draw inspiration from the recently introduced IP-Adapter~\cite{ye2023ip}.  The IP adapter uses a CLIP image encoder to extract image features that can be injected into a diffusion model via the cross-attention layers. These features can be used as an additional condition to guide text prompts or other mediums for the generation. For example, one can input an image of a person and then describe ``on the mountain'' with text to obtain an image of the person in the mountains.

However, we realize that IP-Adaptor does not work well when combined with an Img2Img pipeline, as shown in Figure \ref{fig:ip_adaptor_ablation} (a) for our task. Moreover, adding text guidances like ``changing the apple texture to golden bowl'' does not produce photorealistic output and does not preserve other scene information (\textit{i.e.} background). This problem of geometry and material entanglement within material embedding $z_M$ remains unsolved, thus motivating the need for geometry and illumination guidance.

\subsection{Geometry Guidance via Depth Estimation}
Since decoupling geometry and material properties in images is challenging and requires additional training data, we provide an alternative solution where we enforce a stronger geometry prior to the diffusion model to overwrite the structural information present in $z_{M}$. To this end, we adopt a depth-based ControlNet to provide geometry guidance from the input image $I$. We observe that the geometry information from the depth map $D_I$ overwrites the geometry information encoded in the $z_M$ (see Figure \ref{fig:ip_adaptor_ablation} (b)). Note that with the geometry enforced by using depth-based ControlNet, we can successfully transfer the golden material of the bowl to the apple. 

While the use of ControlNet with IP-Adaptor is introduced in the original IP-Adaptor paper~\cite{ye2023ip}, we employ it for a different purpose of applying new structural control over an object in the image (\textit{e.g.,} changing a person's pose). After extensively comparing various components for encoding the material exemplar and input image (analysis in Section~\ref{sec: qualitative_comparison}), we find the depth-based guidance from pre-trained ControlNet helps us preserve the original geometry of the object for the task of material transfer.

While the addition of ControlNet helps preserve the geometry, we observe that the results suffer from inconsistency in preserving the illumination and background from the input image. This is evident in Figure \ref{fig:ip_adaptor_ablation}, where the background and the lighting changes differ from the input.

\subsection{Latent-space Illumination Guidance}
\label{sec:new_illum}
Our final branch is primarily responsible for preserving the illumination and background in the input image. We propose two-fold guidance for illumination in the latent space during generation -- an inpainting module and a foreground decoloring process. In addition to the attached IP-Adaptor and ControlNet, we adopt an inpainting diffusion model $\mathcal{S}$ instead of a standard generator. Specifically, our ControlNet-inpainting procedure takes in four conditions for image generation:
\begin{equation}
I_{gen} = \mathcal{S}(z_{M}, D_I, I_{init}, F),
\end{equation}
where $z_M$ is the material encoding, $D_I$ is the depth map computed for input image $I$, $I_{init}$ is the initial image to denoise from, and $F$ is the foregound mask of target object in $I$ which we are editing.

\begin{figure}[t]
    \centering
    \includegraphics[width=\columnwidth]{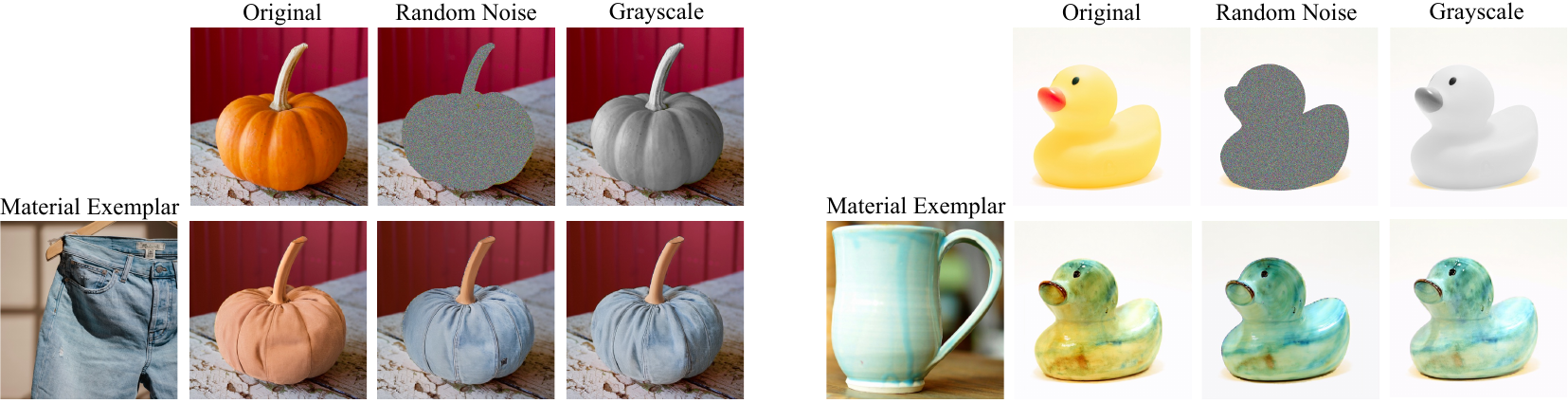}
    \caption{\textbf{Ablating input for illumination guidance.} To validate our design choice of the foreground-grayscale image for initializing inpainting, we compare the generated results against using the original image and random noise as inputs. The original image presents a strong base color prior that perturbs the generation, while the random image neglects shading information, leading to wrong lighting in both examples.}
    \vspace{-1.5em}
    \label{fig:basecolor_change}
\end{figure}

We conduct an ablation on the various versions of $I_{init}$, as shown in Figure \ref{fig:basecolor_change}.
Specifically, we test out the following settings: (1) using the original input image, (2) initializing the foreground with random noise, and (3) using the foreground grayscaled image.
Intuitively, directly letting $I_{init} = I$ (Setting (1)) would be a preferable option as $I$ encompasses implicit lighting information (from the object's shading and the surrounding environment) while conveniently enforces all other parts of the image other than the object to remain the same. In practice, however, we found that using the original image inevitably introduces a strong prior of the base color from the input object (e.g. orange color of pumpkin), which would be entangled with the material base color from $M$ in the output image. This artifact is sustained even when we significantly extend the number of denoising steps. On the other hand, when initializing $I_{init}$ with random noise, the method indeed removes the base color prior but also removes the shading information causing incorrect illuminations in the synthesized object (e.g., the left side of the synthesized pumpkin is darker, but light is coming from the left). In our proposed pipeline, we perform grayscale operations in the pixel space for the object region (3). This provides a balanced solution of removing the strong color priors from the input image while keeping the shading cues for the inpainting diffusion model.

Thus, we propose to initialize $I_{init}$ as:
\begin{equation}
I_{init} = F \odot I_{gray} + (1 - F) \odot I,
\label{grayscale}
\end{equation}
which is converting the appearance of foreground object in the image to grayscale. By doing so, $(1 - F) \odot I$ implicitly preserves the lighting direction, intensity, and color information, while $F \odot I_{gray}$ preserves the shading information of the object without any base color prior.

\subsection{Implementation Details}
We implement our method using Stable Diffusion XL Inpainting~\cite{podell2023sdxl} with the corresponding version of depth-based ControlNet~\cite{zhang2023adding} and IP-Adaptor~\cite{ye2023ip}. We use Dense Prediction Transformers for depth estimation \cite{ranftl2021dpt} and Rembg\footnote{\url{https://github.com/danielgatis/rembg}} for foreground extraction. Our method is implemented in PyTorch and runs on a single Nvidia A-10 GPU with 24 GB of RAM. For all Dreambooth approaches, we use the official LoRA-Dreambooth provided by Diffusers.

\section{Experiments}
We evaluate the efficacy of our method against various baselines. We also present several examples of downstream applications using our method.

\begin{figure}[!t]
    \centering
    \includegraphics[width=\textwidth]{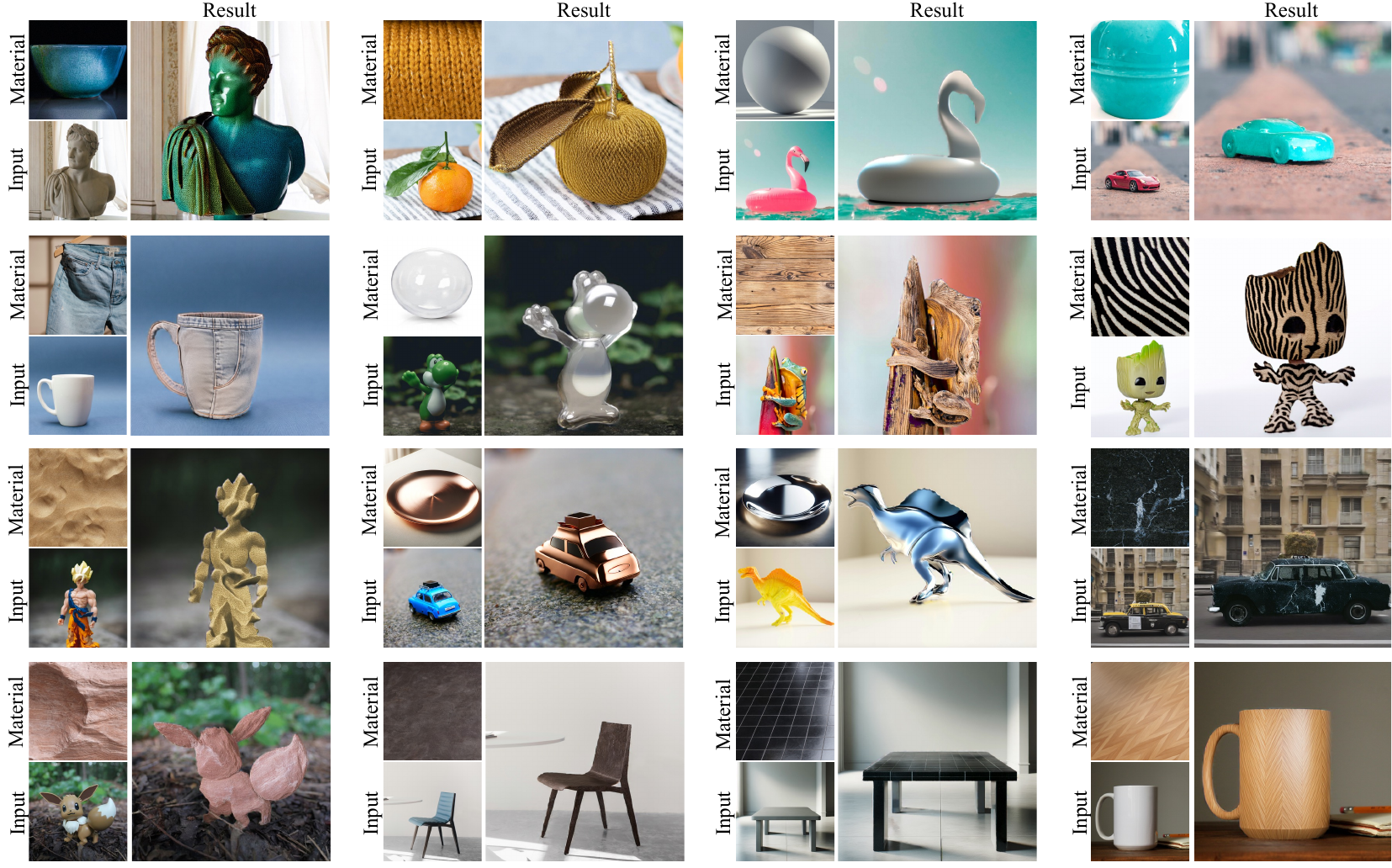}
    \caption{\textbf{Qualitative results on diverse materials.}  We present results of material transfer from a diverse set of material exemplar images. Even when perturbed by lighting and complex geometry, \pipelinename can still isolate the material information from the exemplar image and transfer to various objects while preserving the original geometry and illumination conditions. Note the change in specular regions as shinier materials are chosen in the case of the car made of brass and the dinosaur made of shiny steel.} 
    \vspace{-1em}
    \label{fig:qualitative_results}
\end{figure}
\subsection{Datasets}
As the first to propose this problem, we create two datasets for comparison and evaluation. The real-world datasets provide us an understanding of our model's robustness, while the synthetic dataset is used for standard quantitative metrics.

\noindent\textbf{Real-World Dataset.} We curate a dataset comprising of 30 diverse material exemplars and 30 input images, collected from copyright-free image sources (\textit{i.e.} Unsplash) and images generated by DALLE-3. All of these images are object-centric, where there exists a main object in the foreground to which we are extracting the material from or applying the material onto.

\noindent\textbf{Synthetic Dataset.} To perform quantitative evaluation, we use Blender to create a synthesized dataset of 9 materials randomly initialized by adjusting the base color, metallic, and roughness, and 10 meshes of different categories from Objaverse~\cite{deitke2023objaverse}, generating 90 ground-truth renderings. We render spheres assigned with each material individually and use the rendered image the material exemplar and pre-textured mesh rendering as input for all methods.

While \pipelinename is completely training-free, other methods of learning materials (e.g., Dreambooth) require further fine-tuning for every exemplar given. This makes it infeasible to scale up the two datasets. Both our datasets are of comparable sizes to previous works on finetuning diffusion models~\cite{yeh2024texturedreamer,ruiz2022dreambooth}.

\subsection{Qualitative Results}
\label{sec:qualitative_results}
\noindent\textbf{Material transfer results on real images.}
To demonstrate the application of \pipelinename on a wide range of materials and objects, we present examples of material transfer in Figure \ref{fig:qualitative_results}. The first three rows present results on real-world images, while the fourth row shows results using PBR materials \footnote{\url{https://www.textures.com/browse/pbr-materials/114558}}. Based on the examples, we observe that the material is properly disentangled from the geometry in the material exemplar and follows the shape of the object in the input image. This is particularly evident in the results of the orange, frog, and Groot toy figure, where the material is completely flat. We also notice accurate shadings in the bust and table examples when comparing them against their inputs. In the car and toy dinosaur examples, the reflections from the exemplars are isolated from the textural patterns and cast reasonably based on the illumination cues. 

\noindent\textbf{Qualitative comparisons.} 
\label{sec: qualitative_comparison}
\begin{figure}[!t]
    \centering
    \includegraphics[width=\columnwidth]{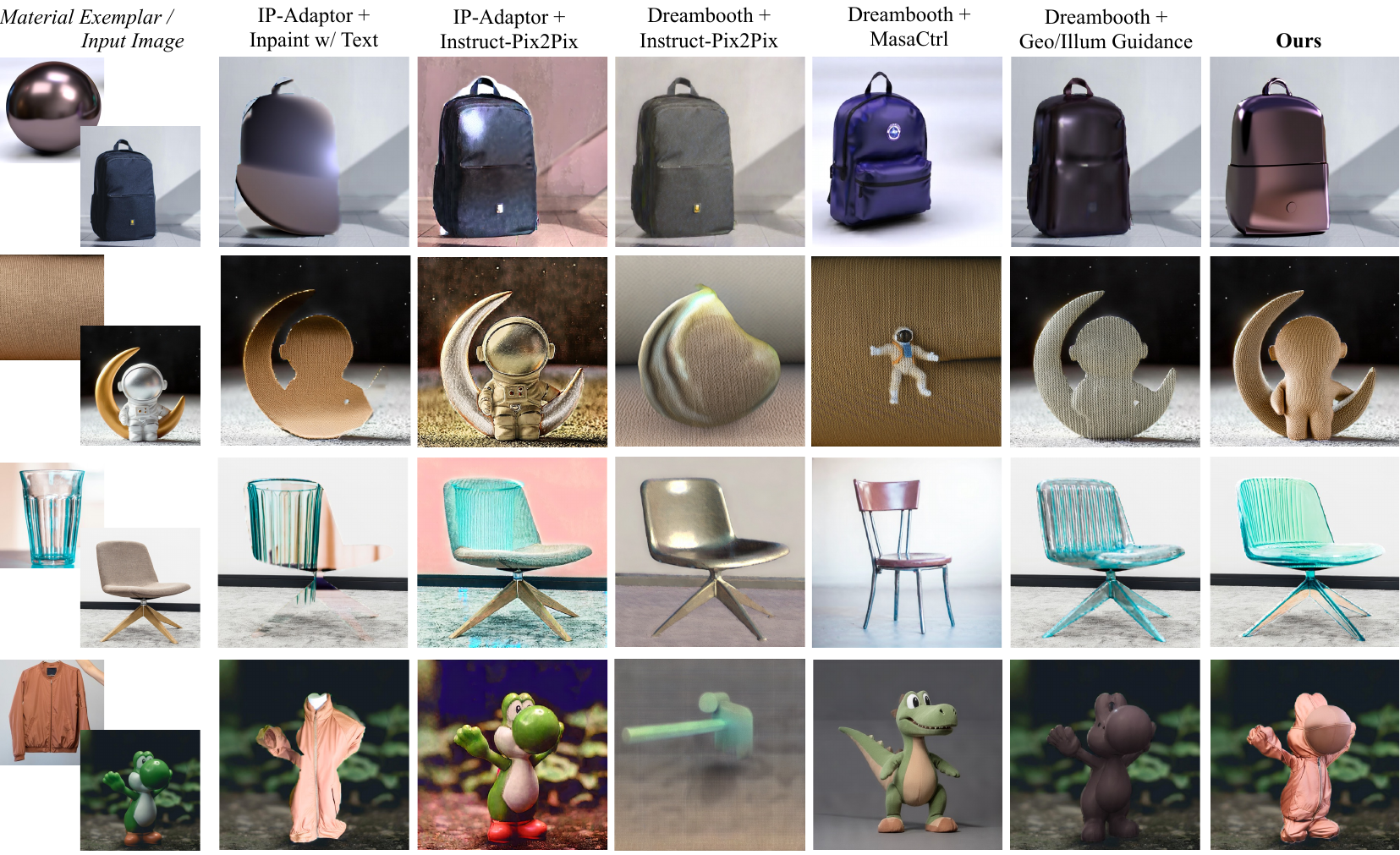}
    \caption{\textbf{Qualitative comparisons against baselines.} Given the material exemplar and input image in the first column, we compare our method to five different baselines. Without any geometry guidance, all image editing baselines fail to impose the correct geometry of the input image. On the other hand, using Dreambooth with our geometry and illumination guidance often contains albedo shifts, potentially due to information loss when encoding material properties into a word token.
    } 
    \vspace{-1em}
    \label{fig:qualitative}
\end{figure} 
Since our work is the first to perform material transfer in latent space, we modified existing methods to compare against. Specifically, since existing image-guided texture synthesis methods utilize Dreambooth for their first step to encode the textures from images into word tokens~\cite{corneanu2024latentpaint,richardson2023texture,yeh2024texturedreamer}, we set Dreambooth as the backbone for learning material properties and combine with text-guided image editing techniques for comparison, including MasaCtrl and Instruct-Pix2Pix, and using \pipelinename but swapping out the IP-Adaptor with text. While our method is training-free, Dreambooth requires finetuning for every material exemplar given. We also explore alternative options to combine with IP-Adaptor, including text-guided inpainting and Instruct-Pix2Pix with the prompt ``Change the texture of the object''.

We present qualitative comparisons against the baselines on four exemplar and input images in Figure \ref{fig:qualitative}. By using Inpainting with Text prompt instead of ControlNet, the model ignores the geometry of the original input when casting the materials. In both cases when using Instruct-Pix2Pix (with IP-Adaptor or Dreambooth), the geometry of all objects is better preserved, but the model fails to capture the material property from the material exemplar image. The combination of Dreambooth and MasaCtrl fails to preserve the geometry of the object in the input image and misattributes the material. The closest baseline to ours is Dreambooth with our proposed geometry and illumination guidance; however, we observe that the word encoding process results in some information loss as evident in the color shifts of the backpack and the astronaut figure. Furthermore, the method requires additional training for every material exemplar, whereas \pipelinename takes roughly 15 seconds to generate the image.

Our method, \pipelinename, performs the task effectively by retaining the object geometry, scene illumination, and attributing the material correctly. Additionally, note that \pipelinename adapts to more challenging material exemplar images, such as transparent materials (glass cup in Figure \ref{fig:qualitative} Row 3) and images with other minor objects (additional hand in Figure \ref{fig:qualitative} Row 4).

\subsection{Quantitative Comparisons} 
We follow previous work \cite{sharma2023alchemist,yeh2024texturedreamer} and use the synthetic images to compare all methods in terms of PSNR, LPIPS~\cite{zhang2018unreasonable}, and CLIP similarity score~\cite{radford2021learning} against ground truth renderings. We grab IP-Adaptor + Instruct-Pix2Pix and Dreambooth + our geometry and illumination guidance as baselines, as they are the strongest (and only) performers from our qualitative comparisons that can roughly edit the material based on the geometry.

\begin{table}[t!]
     \caption{\textbf{Quantitative comparisons and User study}. We grab the strongest baselines in our qualitative comparisons for additional studies. Left: We measure the PSNR, LPIPS~\cite{zhang2018unreasonable}, and CLIP similarity score~\cite{radford2021learning} in a quantitative study on the synthetic dataset. Right: We perform a user study to evaluate the material fidelity and photorealism of the edited images from each method.}
    \centering
    \resizebox{0.48\linewidth}{!}{

    \begin{tabular}{lccc}
    \toprule
    & PSNR$\uparrow$ & LPIPS$\downarrow$ & CLIP$\uparrow$ \\
    \midrule
    IP-Adaptor + Instruct-Pix2Pix  & 16.92 & 0.096 & 0.745\\
    DB + Our Geo/Illum. Guidance  & 25.46& 0.053& 0.893\\
    Ours & \textbf{25.82} & \textbf{0.046}& \textbf{0.899}\\  

    \bottomrule
    \end{tabular}
    }
    \resizebox{0.48\linewidth}{!}{

    \begin{tabular}{lcc}
    \toprule
    &  Fidelity$\uparrow$ & Photorealism$\uparrow$ \\
    \midrule
    IP-Adaptor + Instruct-Pix2Pix  &  1.48& 3.23\\
    DB + Our Geo/Illum. Guidance  & 3.25 & 3.41\\
    Ours & \textbf{4.05} & \textbf{3.78}\\  

    \bottomrule
    \end{tabular}
    }

    \label{tab:quantitative_study}
\end{table}

Table \ref{tab:quantitative_study} (left) presents our results. We see a dramatic improvement when shifting from the instruct-pix2pix pipeline to our geometry and illumination guidance. While using Dreambooth performs similarly to our IP-Adaptor in the synthetic dataset, it requires a fine-tuned model for each material exemplar, making it unfeasible to scale up. In addition, we show in the next section that our method excels in real-world datasets.

\noindent\textbf{User Study.} 
We also create a user study with 16 participants to understand the capability of our model given real-world materials tested on real images. Each subject is shown 5 random samples from the 900 combinations generated from the dataset with our method and against the two strongest baselines: Dreambooth + ControlNet-Inpainting and IP-Adaptor + Instruct-Pix2Pix. We ask each subject to rate each image from 1 to 5 based on (1) material fidelity: how close the material in the generated image is compared to the original exemplar and (2) photorealism: how realistic the generated image is. Our results are summarized in Table \ref{tab:quantitative_study} (right).

Our results show significant improvements from the two baselines in both material fidelity and photorealism of the edited image. The score improvements are also greater in real-world scenarios compared to synthetic ones. This could be the result of information loss during finetuning and overfitting to the exemplar background, which is less significant under controlled synthetic scenarios.

\begin{figure}[t]
    \centering
    \includegraphics[width=\textwidth]{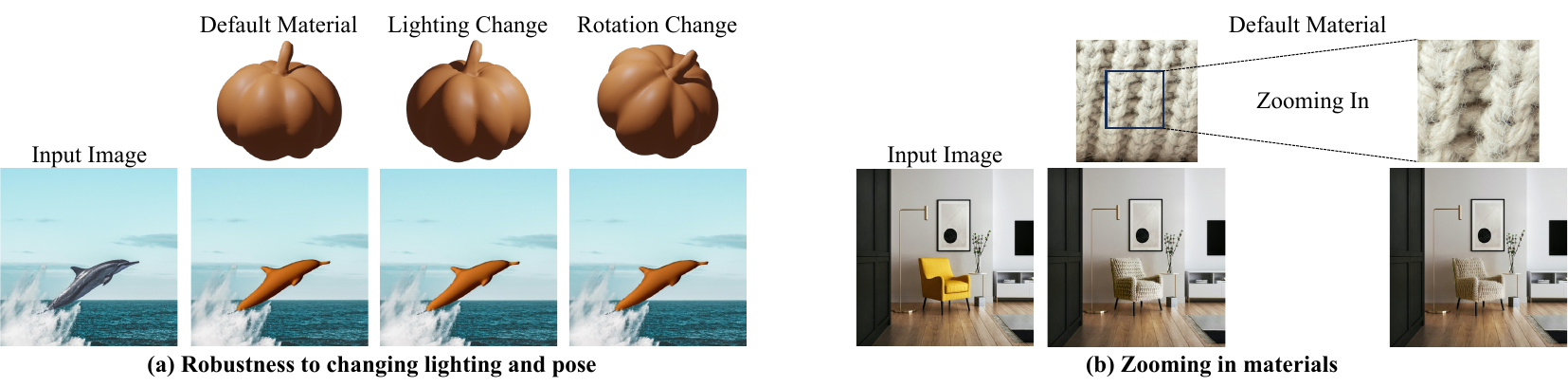}
    \caption{\textbf{Robustness to lighting and object pose.} We present two types of robustness testing. \textbf{(a)}: Robustness to changing the material exemplar lighting and pose. \textbf{(b)}: Zooming into the material exemplar. Our model yields highly similar results in both, showing the capability to adapt to these external changes.
    } 
    \vspace{-2em}
    \label{fig:robustness}
\end{figure}
\subsection{Robustness of the Model}
In addition to the diverse set of results presented in Figure \ref{fig:qualitative_results}, we extensively test out the behavior of \pipelinename with special cases of material exemplar images.

\noindent\textbf{Relighting and rotating the object in the material exemplar image.}
A good material extractor should be agnostic to small lighting and rotation changes of the same object used as the material exemplar. To evaluate this, we render a random material and cast it onto an irregular-shaped pumpkin (another example is in the Appendix). We then render three samples of the pumpkin, a default lighting orientation, a change in lighting direction pitch by 120 degrees, and a random rotation, as shown in \ref{fig:robustness} (a). The transferred materials onto the dolphin remain roughly consistent across all samples, showing that our method is fairly resistant to these changes at a small scale.

\noindent\textbf{Effect of image scale of material exemplar image.}
To examine the effect of the scale of the material exemplar, we first use an image of a woolen cloth material with a distinctive repeating pattern and apply our method to an image of a chair. Then, we zoom into the exemplar image manually to the edge only very few repeated patterns are left. Our results in Figure \ref{fig:robustness} (b) show that while the scale of the material is drastically different, the model automatically re-adjusts the patterns into a reasonable size to be cast onto the input image.

\subsection{Applications}
\noindent\textbf{Applying multiple materials to multiple objects. }
\begin{figure}[t]
    \centering
    \includegraphics[width=\columnwidth]{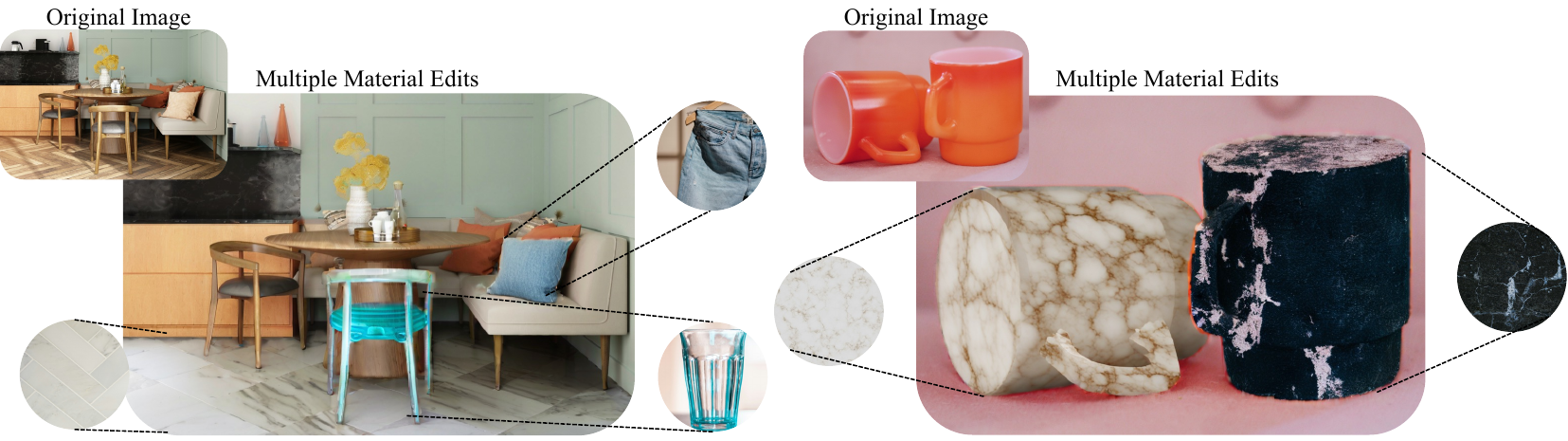}
    \caption{\textbf{Multiple Material Transfers in a Single Image.} By replacing the foreground extraction with an open-vocabulary segmentation module (\textit{e.g.,} SAM) to obtain multiple masks, \pipelinename can be applied iteratively to cast different material properties to different objects in a single RGB image.}
    \label{fig:multi_obj_edit}
\end{figure}

By replacing the foreground extraction with an open-vocabulary segmentation module (\textit{e.g.,} SAM) to obtain multiple masks, \pipelinename can be used to iteratively change multiple materials in a single RGB image. Figure \ref{fig:multi_obj_edit} presents two examples of editing multiple objects in a single image. As evident in the transparent glass chair where the wooden table behind is roughly visible, \pipelinename can generalize complex scenes with multiple objects. Note that the order of multiple object edits also matters. In particular, if the material of one object is reflective of the other, it would be advisable to apply this material at the latest so that the reflections take into account editing changes already made.

\noindent\textbf{Exemplar-based 3D Texturing.}
\begin{figure}[t]
    \centering
    \includegraphics[width=\columnwidth]{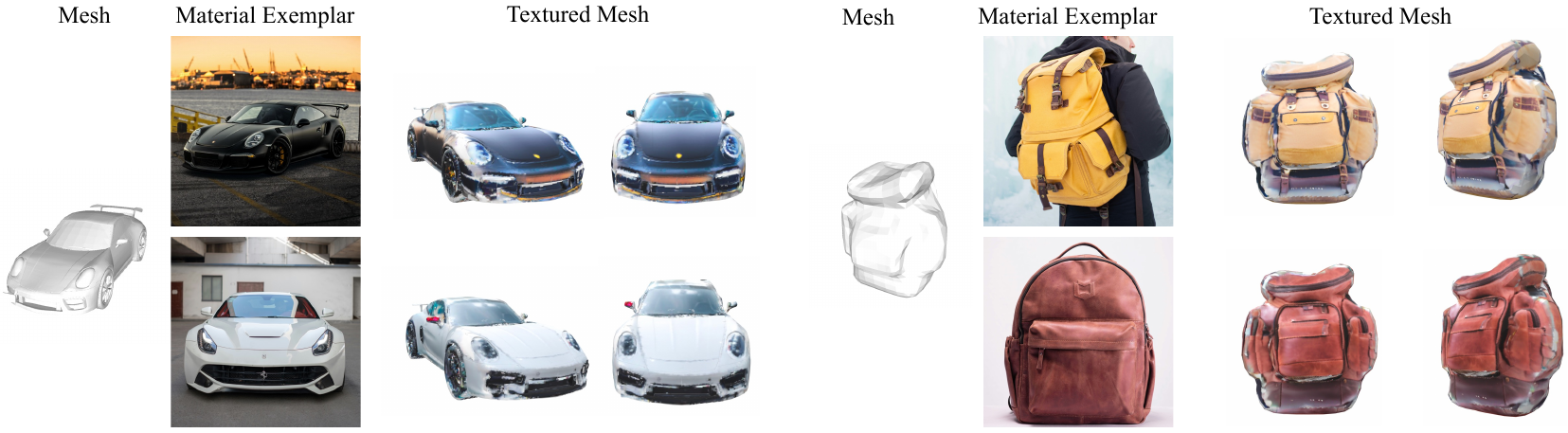}
    \caption{\textbf{Exemplar-prompted 3D Texturing.} We can also combine \pipelinename easily with existing text-driven texturing techniques \cite{chen2023text2tex}. We show examples from two meshes, each using to two material exemplars.
    } 
    \label{fig:application_3dtex}
\end{figure}
\begin{figure}[t]
    \centering
    \includegraphics[width=\columnwidth]{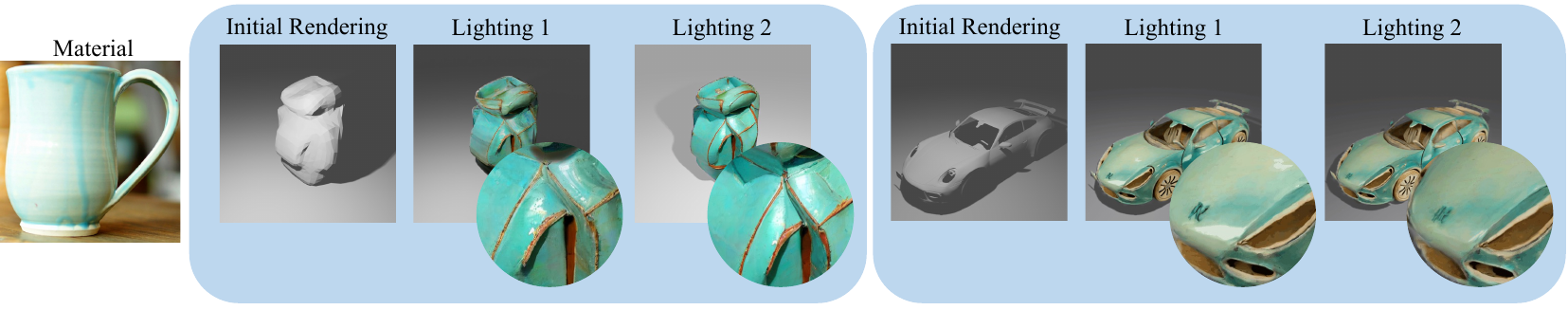}
    \caption{\textbf{Lighting-aware Image Editing.} Given a rendering of a untextured mesh, we can alter \pipelinename slightly to achieve lighting-aware material edit. It can be seen from both examples where the reflection can be disentangled from the object texture. 
    } 
    \vspace{-1em}
    \label{fig:relighting}
\end{figure}
\pipelinename can also be amended to apply textures onto 3D meshes. Recently proposed text-driven texturing pipeline -- Text2Tex \cite{chen2023text2tex} -- uses 2D text-to-image Stable Diffusion and ControlNet to render and refine mesh texture on a per-view basis
with a text prompt. We can replace this texturing backbone of Text2Tex directly with \pipelinename and remove the illumination guidance branch, enabling mesh texturing with a material exemplar image. Figure \ref{fig:application_3dtex} presents four examples of exemplar-based texturing (two from a Porsche mesh and two from a backpack mesh). Even when the exemplar varies from the original geometry (see Row 2 for both cases), the model learns to apply the textures appropriately based on the geometry. We also realize that providing text description (\textit{e.g.,} back view of a backpack) in addition to the material exemplar is particularly helpful in making the texture consistent across all views. Note that existing methods for exemplar-based mesh texturing \cite{richardson2023texture} converts the exemplar image(s) into words via Dreambooth \cite{ruiz2022dreambooth} before texturing. Using \pipelinename makes the texturing process much faster and more scalable.

\noindent\textbf{Lighting-aware Material Transfer.}
\label{sec:relighting}
Given a material exemplar image and an untextured mesh rendered under multiple illumination conditions, \pipelinename can also be used to perform lighting-aware material transfer. Specifically, we first generate the materials and textures of the image under Lighting 1 using \pipelinename. Then, by fixing the same seed during generation and using the generating image given the first lighting as the input to the second, we can enforce consistency in the material and texture generated (details of implementation in Appendix), but change the reflections based on the latent space understanding of the material exemplar. We show examples of transferring the glazed cup material to two mesh renders in Figure \ref{fig:relighting}. \pipelinename successfully disentangles the reflections while keeping most textural patterns consistent between the two images. This technique could potentially be applied jointly with other 3D texture synthesis works~\cite{chen2023text2tex}.

\subsection{Limitations}
\begin{figure}[t]
    \centering
    \includegraphics[width=\columnwidth]{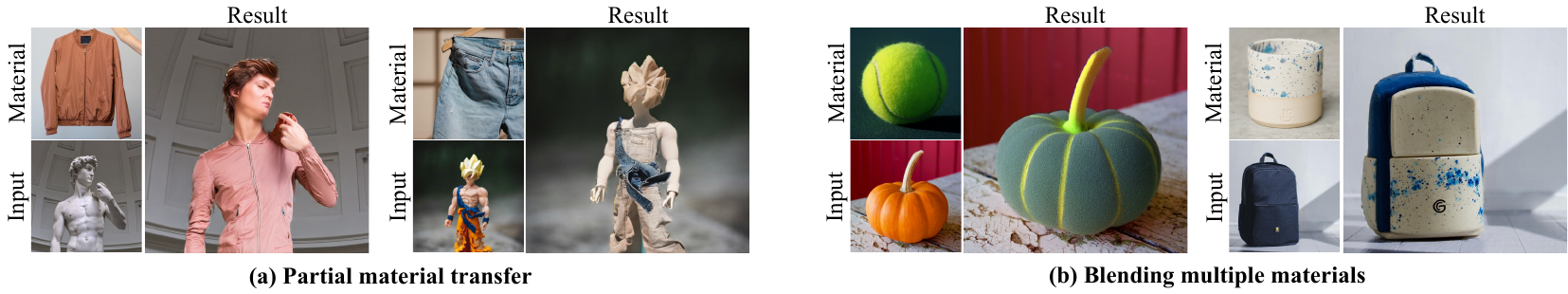}
    \caption{\textbf{Limitations.} Our method primarily fails in two modes. \textbf{(a)} Even when the ControlNet-Inpainting is enforced upon the entire object, the model sometimes picks the most ``probable'' areas to transfer the material, instead of casting the material on the entire object. \textbf{(b)} If two textures are present in the exemplar image (e.g., foreground and background of the tennis ball, the glazed top and bottom logo of the cup), the model sometimes combine both materials when performing the edit.}
    \label{fig:limitations}
    \vspace{-5mm}
\end{figure}

While our method demonstrates generalizable results for the task of single-image exemplar-based material transfer, it still encompasses several limitations. Since we operate majorly in the latent space, the model sometimes exhibits uncontrollable behaviors based on its image understanding. Figure \ref{fig:limitations} presents two forms of more frequent failure cases: (a) Partial material transfer: the material is only transferred to parts instead of the entirety of the object. We hypothesize that the failure stems from the entanglement of material properties and the exemplar's identity, as the material is only applied to where it seems the most probable (\textit{e.g.,} only apply the jacket material to the statue's body). (b) Blending multiple materials: since the current IP-Adaptor does not have a module to extract regions of an image for material transfer, \pipelinename sometimes mixes up multiple materials in the exemplar image during transfer. 

\vspace{-3mm}

\section{Conclusion}

We present \pipelinename, a zero-shot, training-free method for exemplar-based material-editing. \pipelinename is built completely using readily available pre-trained models and demonstrates generalizable and robust results on real images. We curate synthetic and real image datasets to evaluate the performance of our approach. We also demonstrate downstream applications like multiple edits in a single image and material-aware relighting. \pipelinename serves as a strong starting point for future research in image-to-image material transfer, implying opportunities of leveraging pre-trained image diffusion models for complex graphic designing tasks.

%
%
\bibliographystyle{splncs04}
\bibliography{main}
\end{document}